\documentclass[USenglish,twocolumn]{article} \usepackage[T1]{fontenc} 

\usepackage{times} \usepackage[backend=biber,style=numeric,maxbibnames=1,maxcitenames=1]{biblatex}  \usepackage{hyperref}
\usepackage{graphicx}

\title{SANE -- Schema-aware Natural-language Evaluation of Biological Data}
\author{ Rolf Gattung
\thanks{Institute for Automation and Applied Informatics (IAI), Karlsruhe Institute of Technology (KIT), Hermann-von-Helmholtz-Platz 1, 76344 Eggenstein-Leopoldshafen, Germany. Email: rolf.gattung@kit.edu} \and Martin Kr\"uger$^{1}$ \and Markus Reischl$^{1}$ }

\date{} 

\begin{document}

\maketitle

\abstract{
High-throughput microscopy generates large, structured datasets capturing cellular responses to pharmacological perturbations, but accessing these datasets typically requires SQL expertise. Large language models offer a natural-language alternative, yet their tendency to hallucinate raises concerns about result reliability .

We present \textbf{SANE} (Schema-Aware Natural-language Evaluation), a novel paradigm for domain-specific text-to-SQL evaluation: schema-grounded, automatically generated benchmarks tied to real and specific experimental structure. \textbf{SANE} makes evaluation more scalable, systematic, and reproducible.

Using \textbf{SANE}, we evaluate a few-shot large language model and show that, under constrained schemas with structured prompting and guardrails, accurate query generation is achievable without any model training or fine-tuning. Most failures stem from ambiguous or underspecified inputs and manifest as overly cautious clarification requests or answers to queries that should first be disambiguated, rather than incorrect SQL generation. These results indicate that few-shot large language models can provide reliable database access in well-defined domains when combined with schema-aware prompting.
}


\section{Introduction}

High-throughput drug screening using automated microscopy enables systematic analysis of cellular responses across multiple cell lines, compounds, and concentration levels. These experiments generate structured datasets containing both raw measurements and derived analytical values such as EC$_{50}$, drug sensitivity scores, and morphological descriptors.

Platforms such as \textit{Cell-Profiler}~\cite{Carpenter2006} provide infrastructure to process, store, and visualize such datasets. However, effective data access remains a significant challenge, as the available views in such tools are limited and querying the databases requires SQL expertise and familiarity with complex schemas.

Large language models (LLMs) offer a promising solution by enabling natural-language interaction with structured data. Text-to-SQL research has evolved from early neural parsers such as Seq2SQL~\cite{zhong2017seq2sqlgeneratingstructuredqueries} and SQLNet~\cite{xu2017sqlnetgeneratingstructuredqueries} to schema-aware and graph-based models like RAT-SQL~\cite{wang2021ratsqlrelationawareschemaencoding}, PICARD~\cite{scholak2021picardparsingincrementallyconstrained}, and LGESQL~\cite{cao2021lgesqllinegraphenhanced}. Large pretrained models (e.g., T5~\cite{raffel2023exploringlimitstransferlearning}) have further advanced natural language interfaces to structured data. Recent works on robust LLM-based frameworks~\cite{Su2026Text2SQL} and prompting strategies such as DIN-SQL~\cite{pourreza2023dinsqldecomposedincontextlearning} and DAIL-SQL~\cite{gao2023texttosqlempoweredlargelanguage} demonstrate improved zero- and few-shot performance through decomposition and iterative refinement, although models can still hallucinate and make mistakes~\cite{kalai2025languagemodelshallucinate, alansari2026largelanguagemodelshallucination}.

Benchmarks such as WikiSQL~\cite{zhong2017seq2sqlgeneratingstructuredqueries}, Spider~\cite{yu2019spiderlargescalehumanlabeleddataset}, BIRD~\cite{li2023llmservedatabaseinterface}, and conversational extensions like CoSQL~\cite{yu2019cosqlconversationaltexttosqlchallenge} and SParC~\cite{yu2019sparccrossdomainsemanticparsing} evaluate cross-domain generalization, multi-table reasoning, and multi-turn interactions. However, these datasets primarily focus on human-authored queries and general schemas, leaving domain-specific biomedical databases with fixed experimental structures largely untested.

In contrast, we focus on automatically generating schema-grounded evaluation queries directly from the underlying database and experimental design, enabling scalable and reproducible benchmarking tailored to domain-specific biological dataset.
In this work, we evaluate a custom LLM research assistant that queries a PostgreSQL database using a 4-bit quantized LlaMA 3.1 model without any training or fine-tuning. The LLM and database are integrated into a web application planned for public release.
Here, we introduce \textbf{SANE}, a custom framework that systematically derives realistic, schema-grounded test queries from high-throughput drug response databases for rigorous evaluation. Our key contributions are:
\begin{itemize}
\item A scalable framework for domain-specific benchmarking of text-to-SQL systems on high-throughput biological data
\item A large-scale evaluation with 572 non-trivial, automatically generated queries derived from real experiments
\item A detailed analysis of robustness and failure modes, identifying ambiguity in natural language as the primary source of errors
\end{itemize}

\section{System Context}

\begin{figure}
    \centering
    \includegraphics[width=\linewidth]{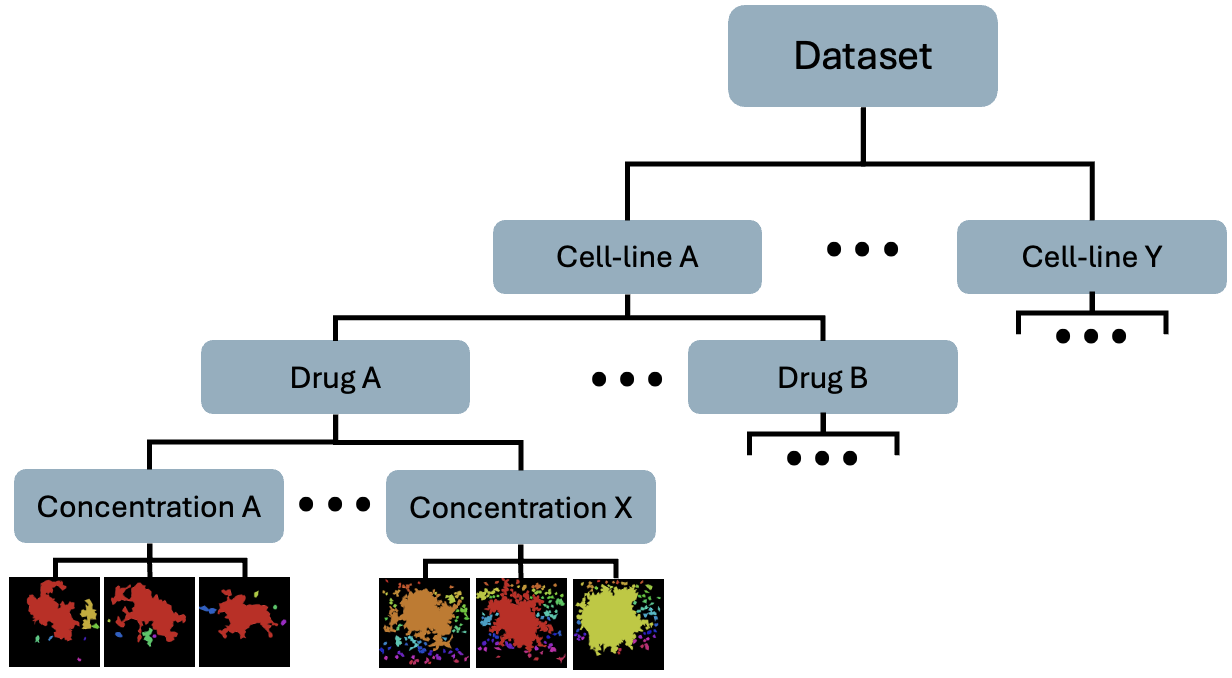}
    \caption{Hierarchical dataset structure of our high-throughput experiments. Each dataset contains multiple cell lines treated with the same drugs at identical concentration ranges. Multiple image replicates are acquired per experimental condition.}
    \label{fig:dataset_vis}
\end{figure}

In high-throughput microscopy, e.g. drug screenings, the datasets typically follow a hierarchical experimental design as in Figure~\ref{fig:dataset_vis}. Experiments include one or more cell lines, each cell-line is tested against multiple drugs at various concentrations. Multiple images of replicates are acquired per condition, reflecting standard drug screening protocols. From these images our webapp extracts morphological features and computes derived statistics (EC$_{50}$, dose–response curves, sensitivity scores), storing them in a relational database.

To query this data we can use a text-to-SQL LLM. In this work, we evaluate the quantized Llama 3.1~\cite{grattafiori2024llama3herdmodels} model running on a single NVIDIA A6000 GPU with 48 GB using virtual large language model(vLLM)~\cite{kwon2023efficientmemorymanagementlarge}. Crucially, the model operates in few-shot mode, no training or fine-tuning is performed. Previous works showed the capabilities of few-shot LLMs \cite{zero-few, brown2020languagemodelsfewshotlearners, info:doi/10.2196/85614}. The LLM interprets natural-language queries about dose–response experiments, generates SQL statements, and retrieves structured data from the underlying relational database. For example, for the question: "How many datasets do I own?" the model would generate a query like "SELECT COUNT(*) FROM datasets d WHERE d.owner\_{id} = user\_{id}" execute it and retrieve the number of datasets of that user.

\begin{figure}
    \centering
    \includegraphics[width=0.5\linewidth]{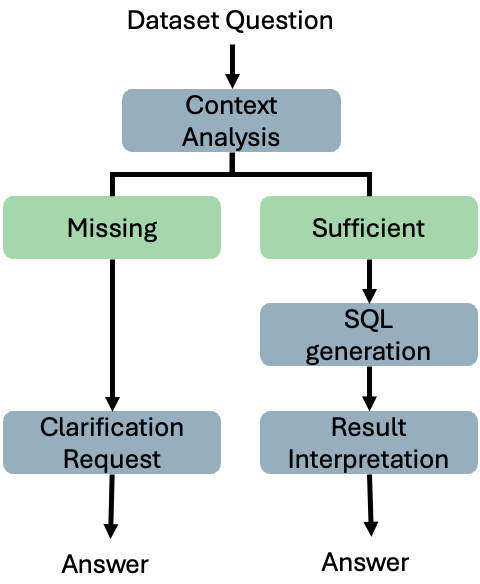}
    \caption{Three stage LLM processing pipeline. First, the model determines whether sufficient context is available to answer the query. If the context is \textit{sufficient}, a SQL query is generated and executed. The result is then interpreted into a natural language response, while \textit{missing} context triggers a clarification request.}
    \label{fig:datset processing}
\end{figure}

The LLM translates natural-language queries into SQL through three stages, visualized in Figure~\ref{fig:datset processing}. First the LLM is tasked to determine whether sufficient context is provided to answer the query, outputting a binary label (\textit{missing}/\textit{sufficient}). If the label is \textit{missing}, a clarification request is returned. Otherwise we generate a SQL statement with the LLM using schema-aware prompting. The returned database result is then interpreted by another LLM pass, turned into a natural language answer, bulletpoint listing or table.

For SQL generation, schema-aware prompting injects database structure, domain terminology, filtering rules (e.g., excluding border artifacts), and dataset-specific context and examples to guide the generation.
The LLM and database are treated as fixed, our focus is systematically evaluating accuracy, robustness, and limitations under realistic scenarios.

\section{SANE}
\begin{figure}
    \centering
    \includegraphics[width=0.5\linewidth]{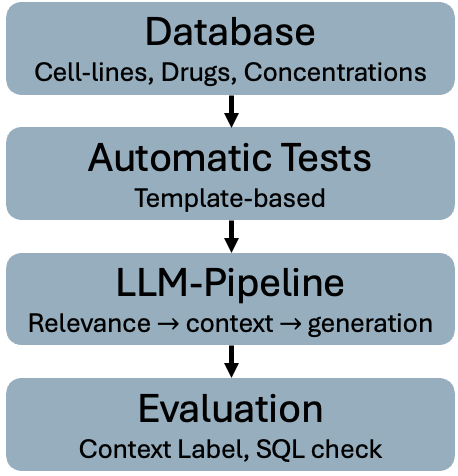}
    \caption{\textbf{SANE} benchmark generation. The database is systematically queried to extract schema structure and experimental content. Test queries are generated with corresponding ground truth SQL. The LLM is evaluated by comparing predicted context classifications (\textit{sufficient} vs.\textit{missing}) and generated SQL execution results against ground truth.}
    \label{fig:benchmark}
\end{figure}

\textbf{SANE} is a framework that systematically generates natural language queries and corresponding ground truth SQL statements directly from database contents and experimental structure (Figure~\ref{fig:benchmark}). The framework queries a pre-existing database from real experiments to extract schema information and representative data instances.

Based on this information, SANE constructs queries spanning multiple complexity levels, including simple retrieval, complex analytical queries, and multi-step interactions. In addition, it introduces controlled perturbations such as typographical errors, abbreviations, and missing context to simulate realistic user interaction scenarios.

The generated queries per category are filled with randomly sampled real data, such as drug names, and are easily extendable. The respective SQL generation is non-trivial, requiring correct handling of hierarchical relationships, domain-specific terminology, and implicit constraints such as dataset scoping.

We define 69 fine-grained query categories and corresponding SQL statements, fully available in the \href{https://bwsyncandshare.kit.edu/s/NadxeGj8PYisQWZ}{supplementary material} together with the benchmark and LLM code. For clarity, we group the questions into six top-level categories:

\textit{Simple} queries involve straightforward retrieval, e.g., "What is the area of HUH7 cells for Alvocidib at $10\mu M$?".

\textit{Complex} queries require joins, aggregations, rankings, or listings, e.g., What are the top 5 drugs by DSS score for MCF7 on the circularity feature?

\textit{Errors} include queries with abbreviations, typographical errors, or inconsistencies, e.g., what is the mean area for HUH7 treated with 9-IN-41 at 0.02 uM?, where the correct drug name is 9-ING-41.

\textit{Contextless} queries are underspecified and return overly large result sets. For example, “What is the EC50 for MCF7 on the area feature?” retrieves values for all drugs and should be further refined first.

\textit{Multi-step} queries represent conversational scenarios with incremental refinement of information, e.g. specifying the drug in a followup message in the previous scenario.

\textit{Schema} queries request structured information across tables, e.g., "How many concentration rows correspond to drug 9-ING-41?"

For each question posed to the LLM, we compare the predicted context label (\textit{missing} or \textit{sufficient}) with ground truth. For \textit{missing} queries, correctness is determined by label agreement. For \textit{sufficient} queries, correctness additionally requires that the generated SQL execution results match the reference results (result set equivalence).

This evaluation jointly assesses the model's ability to interpret query intent and to generate executable SQL consistent with the database schema.

\section{Results \& Discussion}

The evaluation comprises \textbf{572} automatically generated test cases spanning various database schemas and statistical attributes.
Table~\ref{tab:results} summarizes few-shot performance across the categories.

\begin{table}[t]
\caption{Evaluation results grouped by query type.}
\label{tab:results}
\centering
\begin{tabular}{lrrrr}
\hline
Category              & Total & Correct & False & Accuracy (\%) \\
\hline
Simple      & 159   & 159     & 0         & 100.0        \\
Complex     & 316   & 311     & 5         & 98.4         \\
Errors      & 36    & 31      & 5         & 86.1         \\
Contextless & 22    & 18      & 4         & 81.8        \\
Schema      & 32    & 31      & 1         & 96.9         \\
Multi-step  & 7     & 6       & 1         & 85.7         \\
\hline
\textbf{Total} & \textbf{572} & \textbf{556} & \textbf{16} & \textbf{97.2} \\
\hline
\end{tabular}
\end{table}

The few-shot system achieves \textbf{97.2\%} overall accuracy, demonstrating reliable performance across retrieval, complex analysis, and schema-aware tasks without any training or fine-tuning. \textit{Simple}, \textit{Complex} and \textit{Schema} related questions achieve a high accuracy of 100\%, 98.4\% and 96.9\%. In categories \textit{Contextless}, \textit{Errors} and \textit{Multi-step} the LLM performs slightly worse, with 81.8\%, 86.1\% and 85.7\%.
This high accuracy is achieved through domain-specific and example-driven prompting and by limiting the scope of LLM and the task. Without examples (zero-shot), solely relying on simple scheme injection, results drop to 29.9\% overall accuracy.

Among the 16 few-shot failures listed in Table~\ref{tab:results}, 10 involve incorrect \textit{missing} context label prediction, typically from unknown synonyms or misinterpreted underspecified queries. This is likely driven by the model’s primary focus on accurate SQL generation for well-defined queries. When errors or missing context are introduced, the resulting uncertainty leads to inaccurate query generation or misclassification of the context label.
In 5 cases, the generated SQL is slightly incorrect, e.g. omitting drug identifiers in listings or answering overly broad questions instead of asking for clarification.
One question is answered with the generic dismiss answer of the webapp because it is wrongly interpreted as not data-related (Q: What is the average EC50 for Trametinib for the area feature? A: I'm here to help! Feel free to ask about your data or the application.).
However, this analysis suggests that the effective error rate in practical usage for SQL generation is lower than the benchmark suggests, since most failure cases occur due to wrong context label prediction or omitting of names, not because of wrong numerical values. 


The results demonstrate that our few-shot LLM achieves high reliability in querying complex biological databases when combined with schema-aware prompting and domain-specific constraints, validation its usage as research assistant.

Importantly, performance strongly depends on structured prompting and schema knowledge. Without domain-specific terminology and examples, the LLM cannot reliably produce valid queries. This emphasizes the critical role of prompt engineering in deploying LLMs for specialized database interfaces without requiring model training.

Failure analysis indicates future improvements should prioritize interactive query refinement and disambiguation rather than model fine-tuning as the model seems to struggle with missing context and unclear questions rather than accurate text-to-SQL generation as shown in the results. Most errors stem from ambiguous or erroneous user input, suggesting that conversational clarification mechanisms would substantially improve practical usability. Additionally, synonym expansion and domain-specific entity recognition could reduce context classification errors.

This approach significantly lowers barriers to accessing high-throughput biological data through prompting alone. By eliminating both SQL expertise requirements and the need for model training, researchers can directly interrogate experimental results, accelerating hypothesis generation and data-driven discovery.

\section{Conclusion}

We presented a systematic evaluation of a custom domain-specific LLM interface for natural-language querying of biological databases. Our framework \textbf{SANE} automatically generates realistic, schema-grounded test-cases from experimental designs, enabling comprehensive and reproducible assessment.

Achieving 97.2\% accuracy on 572 queries through prompting alone—without any training or fine-tuning—the system demonstrates strong performance. This highlights its potential for real-world use in constrained, domain-specific environments, particularly for reliable extraction of numerical values. This is especially useful for large databases that otherwise require custom scripts or software and do not offer an easy, interactive access to the data, as in the high-throughput cell-image analysis.

Future work could focus on interactive disambiguation mechanisms for the LLM for ambiguous questions and extending the \textbf{SANE} framework to broader biomedical domains, including clinical databases and multi-omics repositories.



%

{\tiny
\printbibliography
}

\end{document}